\documentclass{article}
\usepackage{iclr2027_conference,times} 
\newcommand{\gain}[1]{\textcolor{green!45!black}{(-#1\%)}}
\newcommand{\regress}[1]{\textcolor{red!70!black}{(+#1\%)}}
\newcommand{\unchanged}{\textcolor{gray}{(0\%)}}

\newcommand{\retgain}[1]{\textcolor{green!45!black}{(+#1)}}
\newcommand{\ppgain}[1]{\textcolor{green!45!black}{(+#1\,pp)}}
\newcommand{\pploss}[1]{\textcolor{red!70!black}{(-#1\,pp)}}
\newcommand{\ppsame}{\textcolor{gray}{(0.0\,pp)}}

\usepackage{amsmath,amsfonts,bm}

\def\eqref#1{equation~\ref{#1}}

\def\1{\bm{1}}

\DeclareMathAlphabet{\mathsfit}{\encodingdefault}{\sfdefault}{m}{sl}
\SetMathAlphabet{\mathsfit}{bold}{\encodingdefault}{\sfdefault}{bx}{n}

\usepackage{amsmath,amssymb,amsthm,booktabs}
\usepackage{graphicx}
\usepackage{float,placeins}
\usepackage{hyperref}
\usepackage{url}
\usepackage{algorithm}
\usepackage{algpseudocode}

\usepackage{wrapfig}
\newtheorem{theorem}{Theorem}

\title{Beyond Conservatism: Recoverability-Conditioned Exploration for Model-Based Imitation Learning}
\author{
\begin{tabular}{@{}l@{}}
\textbf{Xuanlin Chen}$^{1,2}$,
\textbf{Ziyue Wang}$^{1,2}$,
\textbf{Xunlan Zhou}$^{1,2}$,
\textbf{Yuan-yih Shang}$^{1,2}$,
\textbf{Qiang Wu}$^{1,2}$,
\textbf{Shenghua Wan}$^{1,2,*}$
\\[4pt]
$^{1}$School of Intelligent Science and Technology, Nanjing University, China
\\
$^{2}$National Key Laboratory for Novel Software Technology,
\\
\hspace{0.5em}School of Artificial Intelligence, Nanjing University, China\\
$^{*}$Correspondence to: \texttt{wanshenghua0and1@gmail.com}
\end{tabular}
}
\iclrfinalcopy

\begin{document}
\maketitle
\lhead{Preprint}

\begin{abstract}
Model-based imitation learning (MBIL) improves real-environment interaction efficiency by optimizing policies on imagined rollouts from a learned world model. However, the gap between model-induced and real-environment occupancies makes policy learning sensitive to model error. Conservative MBIL mitigates model exploitation during policy optimization, but when real-environment interactions are collected by the same conservative policy, uncertain regions around the expert distribution remain insufficiently sampled. Generic uncertainty-driven exploration, on the other hand, may allocate interaction to novel but task-irrelevant dynamics. We propose \textbf{\emph{RE}}coverability-\textbf{\emph{CON}}ditioned Exploration for Model-Based Imitation Learning (\textbf{RECON}). RECON separates conservative policy learning from active data collection by maintaining a main policy for task execution and an explorer for real-environment interaction. The explorer is optimized based on epistemic uncertainty conditioned on recoverability estimated from multi-step main-policy imagination, focusing data collection on unknown states from which the main policy can still return toward expert behavior.  Experiments on locomotion, navigation and manipulation show consistent gains in interaction efficiency, imitation performance, and robustness, indicating that RECON directs real-environment interaction toward recovery regions around the expert distribution that are underexplored by prior methods, and thereby learns a world model better suited for imitation.
\end{abstract}

\section{Introduction}
Imitation learning (IL) provides a natural way to learn complex behaviors from expert demonstrations without specifying a task reward~\citep{hussein2017imitation}. The simplest approach, behavior cloning (BC), directly fits expert state-action pairs with supervised learning. However, BC is trained on states induced by the expert while being deployed on states induced by the learned policy. Small action errors can therefore move the policy outside the demonstration distribution, where subsequent errors compound~\citep{ross2011reduction,seo2024mitigating}.

Interactive imitation methods address this distribution shift by incorporating learner-induced data. DAgger~\citep{ross2011reduction} repeatedly queries the expert on states visited by the current policy. A different line of work builds on inverse reinforcement learning (IRL), where expert behavior is used to infer a reward that can subsequently be optimized with reinforcement learning~\citep{abbeel2004apprenticeship, finn2016guided}. Adversarial imitation learning (AIL), such as GAIL~\citep{ho2016generative}, avoids explicitly recovering a reward function by training a discriminator to distinguish expert and learner transitions and using its output as a surrogate reward. Because policy optimization is performed on learner-induced rollouts, these methods can correct deviations from the expert distribution, but typically require substantial real-environment interaction, which can be costly, time-consuming, and potentially unsafe in real-world systems~\citep{dulac2019challenges}.

Model-based imitation learning (MBIL) improves this interaction efficiency by learning predictive dynamics and replacing many real rollouts with model-based optimization or imagined rollouts~\citep{englert2013model,rafailov2021visual}. V-MAIL~\citep{rafailov2021visual} learns a latent world model from demonstrations and environment interaction, and performs 
\begin{wrapfigure}{r}{0.54\textwidth}
    
    \centering
    \includegraphics[width=0.53\textwidth]{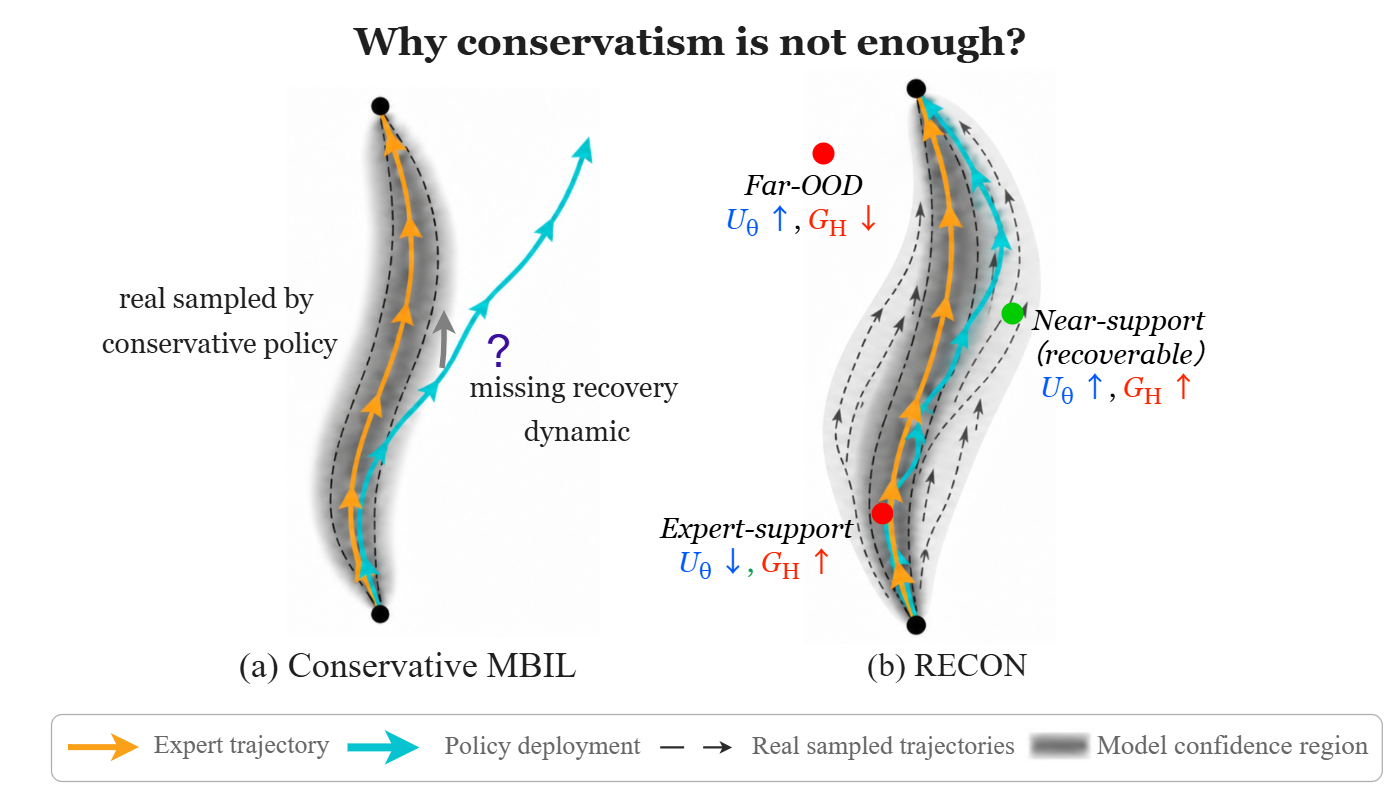}
    \caption{
    \textbf{Conservative model use is not sufficient for model acquisition.}
    A conservative main policy avoids uncertain dynamics, leaving recovery-relevant
    regions underexplored. RECON uses a separate explorer to collect uncertain
    but recoverable transitions.
    }
    \label{fig:overview}
    
\end{wrapfigure}
adversarial policy optimization on imagined on-policy rollouts, while real interaction is used primarily to improve the dynamics model. The learned model, however, introduces another source of error: 
prediction errors can accumulate over imagined rollouts and be exploited by the policy, and the occupancy induced by the learned dynamics can differ from that induced by the real environment~\citep{janner2019trust,ma2023learning}.

This problem is closely related to model exploitation in model-based RL. Conservative model-based methods such as MOPO~\citep{yu2020mopo} and MOReL~\citep{kidambi2020morel} account for model error when optimizing policies from limited data, typically by penalizing or avoiding uncertain model predictions. CMIL~\citep{kolev2024efficient} brings this principle to adversarial imitation learning by augmenting the discriminator-derived imitation reward with an ensemble-based epistemic uncertainty penalty during policy optimization. This naturally raises the \emph{model usage} question:

\begin{quote}
\textbf{\emph{Usage question:}} \emph{Given the current world model, which predictions can be safely used for policy optimization?}
\end{quote}

Conservative optimization reduces the risk of exploiting an inaccurate model, but does not actively improve the model outside its current reliable region. When real-environment data are collected by the same conservative policy, interaction is concentrated in regions that are already well modeled, while uncertain dynamics around the expert distribution may remain insufficiently sampled~\citep{mark2023offline,chen2025offline}. This matters because the model-induced and real-environment occupancies need not coincide: action noise, observation error, or unmodeled disturbances can move the deployed policy into states for which the world model has little data support. Conservative MBIL therefore leaves a complementary \emph{model acquisition} question:

\begin{quote}
\textbf{\emph{Acquisition question:}} \emph{Which unknown dynamics should be prioritized under a limited budget of real-environment interaction?}
\end{quote}

Model-based exploration provides a natural starting point. MAX~\citep{shyam2019model} and Plan2Explore~\citep{sekar2020planning} use information gain or model disagreement to actively collect informative transitions. Their objective is intentionally task-agnostic, aiming to learn broadly useful dynamics models that can support unknown downstream tasks. In imitation learning, however, expert demonstrations already provide a strong task prior~\citep{ren2024hybrid}. Under a limited interaction budget, highly uncertain states far from expert behavior may improve global model accuracy while providing little benefit to the imitation policy~\citep{ma2023learning,aoyama2025poke}.

We therefore focus exploration on states that are both uncertain and \emph{recoverable}. A state is recoverable when the current main policy can still return from it toward expert-compatible behavior~\citep{park2022robust,ankile2024juicer}. Epistemic uncertainty measures the information value of acquiring its dynamics, while recoverability measures its relevance to the imitation task.

Based on this principle, we propose \textbf{Recoverability-Conditioned Exploration for Model-Based Imitation Learning (RECON)}. RECON uses a conservative main policy for task execution and trains a separate explorer solely for real-environment data collection. Recoverability is estimated through multi-step rollouts of the main policy in the Dreamer world model, using the imitation discriminator to evaluate whether the imagined trajectory returns toward expert-compatible behavior. The explorer prioritizes states with both high epistemic uncertainty and high recoverability. The two policies share the world model, representation, discriminator, and replay buffer, allowing exploratory interactions to improve the model used by the main policy without introducing the exploration objective at deployment.

Our contributions are threefold. (a) we distinguish \emph{model usage} from \emph{model acquisition} in conservative model-based imitation and formalize the resulting blind spot in recovery-relevant dynamics. (b) we introduce recoverability-conditioned exploration, using multi-step Dreamer imagination to direct an independent explorer toward uncertain but task-relevant states. (c) under matched interaction budgets, we show on DMC and MetaWorld that RECON improves imitation performance, interaction efficiency, and robustness, and learns a world model with better coverage of recovery-relevant dynamics.

\section{Related Work}

\subsection{Reinforcement and imitation learning}
Reinforcement learning learns control policies by maximizing a task reward
through environment interaction, whereas inverse reinforcement learning infers
a reward function from expert behavior~\citep{abbeel2004apprenticeship,finn2016guided,fu2017learning}.
Imitation learning instead seeks to reproduce expert behavior directly. BC is
simple and interaction-free but suffers from covariate shift. Interactive
methods such as DAgger~\citep{ross2011reduction} obtain corrective supervision on
learner-induced states, while DART~\citep{laskey2017dart} perturbs demonstrations to
expose the learner to nearby deviations. Adversarial methods such as
GAIL~\citep{ho2016generative} formulate imitation as occupancy matching and optimize a
discriminator-derived reward with RL; off-policy variants such as
DAC~\citep{kostrikov2018discriminator} improve data reuse but still depend on
real-environment rollouts. Our work follows this occupancy-matching view but
focuses on improving the dynamics model used for interaction-efficient policy
learning.

\subsection{World models and conservative model-based RL}
Model-based RL learns environment dynamics and uses the model for planning or
policy optimization~\citep{hafner2019learning}. Modern latent world-model methods, including
Dreamer~\citep{hafner2019dream,hafner2020mastering,hafner2023mastering} and TD-MPC \citep{hansen2022temporal,hansen2024td}, learn compact recurrent state
representations and train policies on imagined trajectories. Such model-based
optimization is sample efficient but is vulnerable to model exploitation when
the policy visits regions poorly supported by data~\citep{janner2019trust,yu2020mopo}.

This issue has been studied extensively in offline and data-limited model-based
RL. MOReL~\citep{kidambi2020morel} discourages policies from entering uncertain
regions, while MOPO~\citep{yu2020mopo} optimizes an uncertainty-penalized model MDP.
COMBO~\citep{yu2021combo} similarly introduces conservatism into model-based offline
policy optimization. These methods share the principle that policy optimization
should account for epistemic model error rather than treating all model
predictions as equally reliable. RECON adopts the same principle for the
deployed policy, but addresses a complementary problem: how online interaction
should be allocated to improve the model itself.

\subsection{Model-based imitation learning and active model acquisition}
Model-based imitation combines the sample efficiency of world models with
learning from demonstrations. Earlier approaches combine learned dynamics with
adversarial or trajectory-based imitation~\citep{englert2013model,baram2016model}, while V-MAIL~\citep{rafailov2021visual}
uses a variational latent model to generate approximate on-policy rollouts for
adversarial imitation from high-dimensional observations. CMIL
~\citep{kolev2024efficient} extends this framework with conservative policy optimization,
using ensemble uncertainty to reduce model exploitation.
Other recent methods explore complementary aspects of MBIL~\citep{kidambi2021mobile,hu2022model,zhang2023action}.
DITTO~\citep{demoss2023ditto} performs fully offline imitation through latent
trajectory matching, SeMAIL~\citep{wan2023semail} separates task-relevant dynamics
from visual distractors, and IQ-MPC~\citep{li2025reward} combines reward-free
world-model learning with inverse soft-$Q$ learning and latent model predictive
control. These methods primarily modify the imitation objective, representation,
or controller. RECON instead studies the data distribution used to learn the
world model under a limited real-interaction budget.

Model-based exploration provides a natural acquisition mechanism: MAX
~\citep{shyam2019model} connects ensemble disagreement to information gain, while
Plan2Explore~\citep{sekar2020planning} seeks expected future novelty through latent
imagination. Whereas these methods pursue broad, task-agnostic coverage, RECON
conditions epistemic uncertainty on the task information in expert
demonstrations and prioritizes dynamics relevant to imitation. Related recovery
methods use corrective demonstrations or backward and reverse-model
augmentation to improve behavior outside nominal expert trajectories
~\citep{laskey2017dart,park2022robust,shao2024offline}. RECON instead uses recoverability predicted in imagination, requiring neither a reverse dynamics model nor
synthetic policy labels.

\section{Problem Formulation and Analysis}
\label{sec:analysis}
Let $\mathcal M=(\mathcal S,\mathcal A,T,\mu_0,\gamma)$ be the true MDP and
$\widehat{\mathcal M}_{\mathcal D}$ the model learned from real transitions
$\mathcal D$.  For analysis only, assume a bounded latent task reward
$|r(s,a)|\le R_{\max}$ under which the expert $\pi_E$ is optimal; this reward is
never observed by the learner.  For initial distribution $q$, define the
normalized occupancy $\rho_{\mathcal M,q}^{\pi}$ and one-step error
\begin{equation}
\rho_{\mathcal M,q}^{\pi}(s,a)=(1-\gamma)\sum_{t\ge0}\gamma^t
\Pr(s_t=s,a_t=a\mid s_0\sim q),\qquad
e_{\mathcal D}(s,a)=D_{\rm TV}(T,\widehat T_{\mathcal D})(s,a).
\label{eq:occupancy_model_error}
\end{equation}
\vspace{-5mm}
\subsection{The acquisition blind spot}
The standard simulation argument gives a bound whose model term is evaluated
under \emph{real}, rather than imagined, visitation.
\begin{theorem}[Real-environment model error]
\label{thm:real_model_error}
Let $\varepsilon_q(\pi,\mathcal D)=
\mathbb E_{\rho_{\mathcal M,q}^{\pi}}[e_{\mathcal D}]$. Then
\begin{equation}
|V_{\mathcal M,q}^{\pi}-V_{\widehat{\mathcal M}_{\mathcal D},q}^{\pi}|
\le\frac{2\gamma R_{\max}}{(1-\gamma)^2}\varepsilon_q(\pi,\mathcal D),
\label{eq:real_model_error_bound}
\end{equation}
and
\begin{equation}
V_{\mathcal M}^{\pi_E}-V_{\mathcal M}^{\pi}
\le \frac{2R_{\max}}{1-\gamma}D_{\rm TV}
(\rho_E,\rho_{\widehat{\mathcal M}_{\mathcal D}}^{\pi})
+\frac{2\gamma R_{\max}}{(1-\gamma)^2}\varepsilon_{\mu_0}(\pi,\mathcal D).
\label{eq:imitation_real_error_bound}
\end{equation}
For $d_0=\rho_{\mathcal M,\mu_0}^{\pi_m}$ and
$d_q=\rho_{\mathcal M,q}^{\pi_m}$, if $d_q\ll d_0$ with
$\kappa_q:=\|\mathrm d d_q/\mathrm d d_0\|_\infty<\infty$, then
$\varepsilon_q\le\kappa_q\varepsilon_{\mu_0}$; without this bounded
density-ratio condition, no finite distribution-free factor is guaranteed.
\end{theorem}
Thus conservatism controls which model predictions are used, but nominal
collection controls recovery error only where the recovery occupancy is
covered.  Proofs of all results are in Appendix~\ref{app:proofs}.

\subsection{Recovery-weighted acquisition}
At acquisition round $k$, freeze the current main policy and an
expert-compatibility score $c_k\in[0,1]$.  The $H$-step recoverability of a
candidate state and its transition-level gate are
\begin{equation}
G_{k,H}^{\pi_m}(s)=\frac{1}{C_H}\mathbb E_{\widehat{\mathcal M}_{\mathcal D_k},\pi_m}
\!\left[\sum_{h=1}^H\gamma^{h-1}c_k(s_h,a_h)\mid s_0=s\right],\quad
g_k(s,a)=\mathbb E_{\widehat T_{\mathcal D_k}}[G_{k,H}^{\pi_m}(s')],
\label{eq:round_recoverability}
\end{equation}
where $C_H=\sum_{h=1}^H\gamma^{h-1}$.  Let $T_\omega$ index posterior-plausible
dynamics and define posterior predictive KL
$\ell_{\mathcal D}(x)=\mathbb E_{\omega\mid\mathcal D}
D_{\rm KL}(T_\omega(\cdot\mid x)\Vert\widehat T_{\mathcal D}(\cdot\mid x))$.
For a reference distribution $\nu$ over candidate acquisition inputs, keep
$g_k$ fixed within round $k$ and set
\begin{equation}
\mathcal R_k(\mathcal D)
=
\mathbb E_{x\sim\nu}
[g_k(x)\ell_{\mathcal D}(x)].
\label{eq:recovery_weighted_risk}
\end{equation}
Here $\nu$ serves as a reference measure over transitions. In the practical algorithm, its empirical analogue
is the replay-seeded imagined transition distribution visited by the explorer.
\vspace{+2mm}
\begin{theorem}[Recovery-weighted control]
\label{thm:recovery_weighted_control}
If $\rho_{T_\omega,q}^{\pi_m}(x)\le C_q\nu(x)$ almost surely, then, with
$Z_k=\mathbb E_\nu[g_k]$ and
$\bar e_k(x)=\mathbb E_{\omega\mid\mathcal D_k}
D_{\rm TV}(T_\omega,\widehat T_{\mathcal D_k})(x)$,
\begin{equation}
\mathbb E_{\omega\mid\mathcal D_k}[\varepsilon_q(\omega,\mathcal D_k)]
\le C_q\!\left[\sqrt{Z_k\mathcal R_k(\mathcal D_k)/2}
+\mathbb E_\nu[(1-g_k)\bar e_k]\right].
\label{eq:recovery_weighted_control}
\end{equation}
\end{theorem}
The first term is precisely the part of predictive risk that can affect states
judged recoverable by the current policy; the residual makes explicit that the
criterion does not seek an everywhere-accurate model.
For a real query $x$ with successor $Y_x$, let
$\mathcal D'_{k,x}=\mathcal D_k\cup\{(x,Y_x)\}$ and let $Y_z$ be an independent
successor conditional on $\omega$.  Bayesian risk reduction yields:
\begin{theorem}[Recovery-weighted information gain]
\label{thm:recovery_information_gain}
\begin{equation}
\mathcal R_k(\mathcal D_k)-\mathbb E_{Y_x\mid\mathcal D_k}
[\mathcal R_k(\mathcal D'_{k,x})]
=\mathbb E_{z\sim\nu}[g_k(z)I(Y_z;Y_x\mid z,x,\mathcal D_k)].
\label{eq:recovery_information_gain}
\end{equation}
For a discrete, posterior-independent cell model whose successors are
deterministic given $\omega$, this reduces to
$\Gamma_k(x)=\nu(x)g_k(x)I(\omega;Y_x\mid x,\mathcal D_k)$.
\end{theorem}
Approximating information gain by ensemble disagreement $U_\theta$
gives the complementary objectives
\begin{equation}
J_m=J_{\rm IL}(\pi_m)-\alpha\mathbb E_{\rho_{\widehat{\mathcal M}}^{\pi_m}}[U_\theta(s,a)],
\qquad
J_e=J_{\rm IL}(\pi_e)+\beta\mathbb E_{\rho_{\widehat{\mathcal M}}^{\pi_e}}
[U_\theta(s,a)G_{k,H}^{\pi_m}(s')].
\label{eq:final_objectives}
\end{equation}
The negative sign protects policy optimization; the positive, gated term
decides which reachable uncertainty is worth resolving with real interaction.

\begin{figure}[t]
  \centering
  \includegraphics[width=1\linewidth]{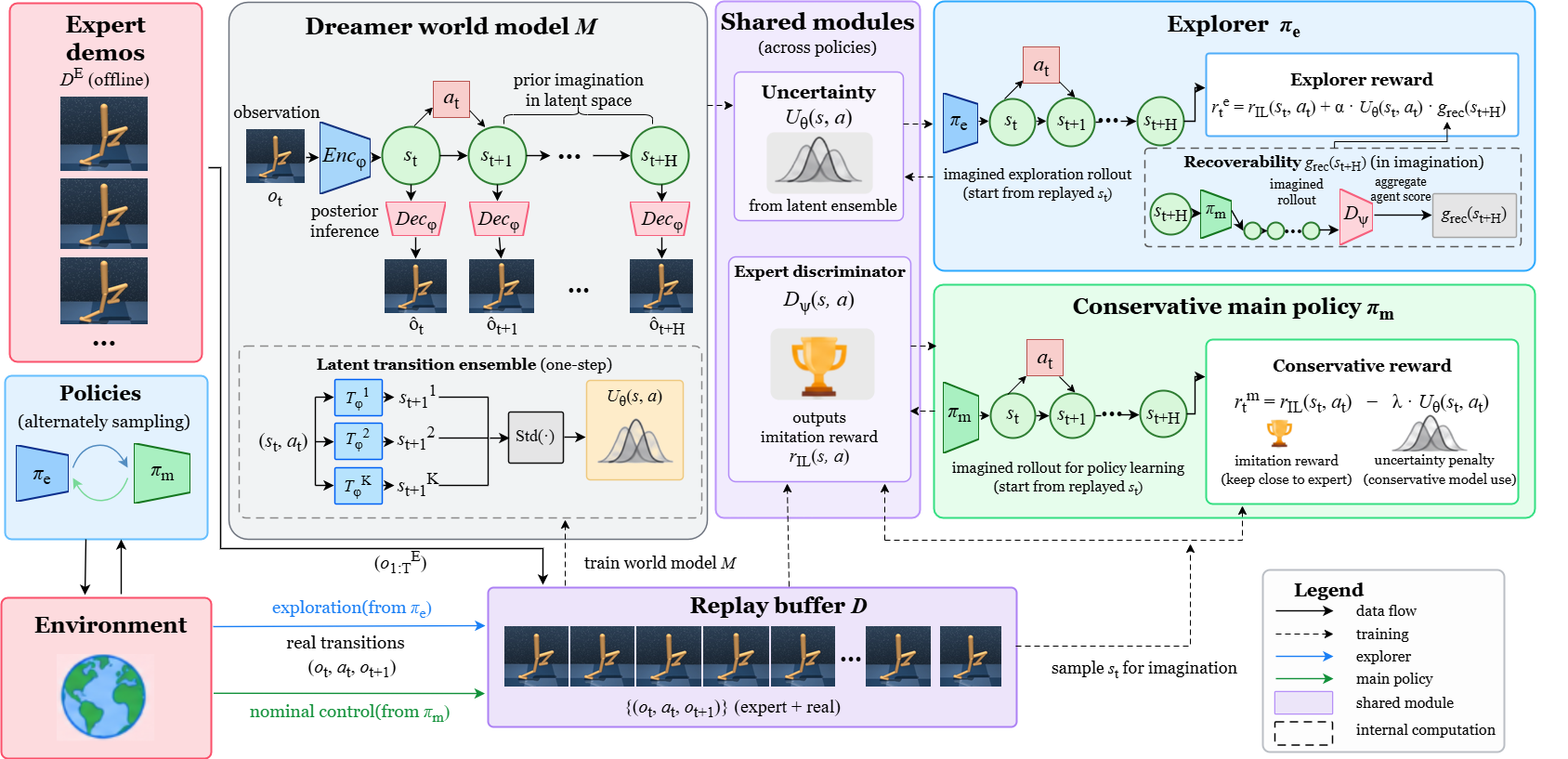}
  \caption{\textbf{RECON architecture on the DreamerV2 backbone}. The conservative main policy uses uncertainty
  pessimistically, while a separate explorer uses recoverability-gated
  uncertainty for acquisition. Both update one world model; only the main policy
  is deployed.}
  \label{fig:architecture}
\end{figure}

\section{RECON: Recoverability-Conditioned Exploration}
\label{sec:method}
RECON extends CMIL's DreamerV2-based latent adversarial-learning loop~\citep{kolev2024efficient,hafner2020mastering} with a
separate recoverability-conditioned explorer, while sharing the representation,
world model, discriminator, and replay.
(Figure~\ref{fig:architecture}).

\paragraph{Latent world model.}
We use DreamerV2 with the continuous ensemble RSSM configuration
of CMIL. Let $s_t=(h_t,z_t)$ denote its deterministic and Gaussian stochastic
state,
\begin{equation}
 h_t=f_\theta(h_{t-1},z_{t-1},a_{t-1}),\qquad
 z_t\sim q_\theta(z_t\mid h_t,o_t),\quad
 \widehat z_t\sim p_\theta(z_t\mid h_t),\quad
 o_t\sim p_\theta(o_t\mid s_t).
 \label{eq:rssm}
\end{equation}
Posterior trajectories from expert and online replay train the model with a
reconstruction--KL objective (using KL balancing and free nats in practice),
\begin{equation}
 \mathcal L_{\rm wm}(\theta)=
 \mathbb E_{\tau\sim\mathcal D_E\cup\mathcal D}
 \sum_t\left[-\log p_\theta(o_t\mid s_t)
 +c_{\rm KL}D_{\rm KL}\!\left(q_\theta(z_t\mid h_t,o_t)
 \Vert p_\theta(z_t\mid h_t)\right)\right].
 \label{eq:world_model_loss}
\end{equation}
Policy rollouts instead use the recurrent prior. A bootstrap transition ensemble~\citep{chua2018deep,lakshminarayanan2017simple}
predicts means $\mu_k(s_t,a_t)$, giving the epistemic score
\begin{equation}
 U_\theta(s_t,a_t)=\frac{1}{d_z}\sum_{j=1}^{d_z}
 \operatorname{Std}_{k=1:K}[\mu_{k,j}(s_t,a_t)].
 \label{eq:implemented_uncertainty}
\end{equation}

\paragraph{Adversarial imitation in imagination.}
Starting from replay posteriors, policy $i\in\{main,explore\}$ generates
$\widehat a_t^i\sim\pi_i(\cdot\mid\widehat s_t^i)$ and
$\widehat s_{t+1}^i\sim p_\theta(\cdot\mid\widehat s_t^i,\widehat a_t^i)$.
As in CMIL, the discriminator contrasts expert posterior transitions with
on-policy imagination:
\begin{equation}
 \mathcal L_D(\psi)=-\mathbb E_{(s,a)\sim\rho_E}\log D_\psi(s,a)
 -\mathbb E_{(s,a)\sim\widehat\rho_{\pi_m}}\log(1-D_\psi(s,a)).
 \label{eq:discriminator_loss}
\end{equation}
The discriminator therefore remains the common imitation signal; uncertainty
changes how each policy uses the learned model.

\paragraph{Dual policy objectives.}
Both policies use the same DreamerV2 latent actor--critic and expert-action
regularizer with the shared optimization specified in
Appendix~\ref{app:dreamer_backbone}, but have separate actors and critic ensembles. The acquisition distinction
is entirely expressed by their imagined rewards,
\begin{equation}
 r_t^m=D_\psi(\widehat s_t^m,\widehat a_t^m)-\alpha U_\theta(\widehat s_t^m,\widehat a_t^m),
 \qquad
 r_t^e=D_\psi(\widehat s_t^e,\widehat a_t^e)+\beta U_\theta(\widehat s_t^e,\widehat a_t^e)G_T.
 \label{eq:dual_rewards}
\end{equation}
With $\alpha=\beta=10$, $\pi_m$ retains CMIL's pessimistic model-use objective,
whereas $\pi_e$ assigns optimistic value only to uncertainty conditioned on recoverability.

\paragraph{Recoverability-conditioned acquisition.}
To compute recoverability $G_T$, rather than learning a separate
inverse or backward dynamics model, RECON exploits Dreamer's multi-step latent
imagination to directly evaluate whether the current main policy can return
from the explorer endpoint toward expert-compatible behavior. RECON takes the terminal state $\widehat s_T^e$ of an explorer
imagination and rolls the frozen main policy forward deterministically for
$H_r$ steps,
\begin{equation}
 \widetilde a_h=\operatorname{mode}\pi_m(\cdot\mid\widetilde s_h),\quad
 \widetilde s_{h+1}=p_\theta(\cdot\mid\widetilde s_h,\widetilde a_h),\quad
 G_T=\max_{h=H_r-L+1:H_r}
 \sigma\!\left(\frac{D_\psi(\widetilde s_h,\widetilde a_h)-\delta_k}{\tau}\right).
 \label{eq:implemented_gate}
\end{equation}
Here $\delta_k$ is the exponential moving average of the main-policy
discriminator score,  $\sigma(x)$ is the logistic sigmoid. The tail maximum asks whether the current
main policy can regain main-compatible behavior, rather than whether the
explorer endpoint merely looks expert-like. A single terminal-derived $G_T$ is shared by all transitions of an imagined
explorer rollout.  During the explorer update, model,
discriminator, and main-policy parameters are frozen, but the pathwise gradient
through the recovery rollout is retained (detailed in Appendix~\ref{app:gradient}). We use $H_r=5$, $L=3$, and
$\tau=0.05$.

\paragraph{Real interaction.}
The two policies alternate \emph{within} each real episode rather than using
separate rollouts. Each episode is partitioned into 15 temporal strata, and one
10-step explorer window is sampled uniformly inside every stratum; the main
policy acts at all remaining steps. Thus, a full 500-step task episode contains 150 explorer and 350
main-policy steps. All transitions enter the same replay, which updates the
shared DreamerV2 model, discriminator, and both actor--critics. Evaluation
removes the explorer and executes only the deterministic main policy. The complete collection and training loop is given in
Appendix~\ref{app:algorithm}, with shared hyperparameters in
Appendix~\ref{app:hyperparameters}.

\section{Experiments}
\label{sec:experiments}
We test the acquisition claim through five questions: \textbf{(a)} Does RECON
improve imitation at a fixed interaction budget? \textbf{(b)} Does targeted acquisition improve robustness beyond nominal training? \textbf{(c)} Where does
$G$ redirect interaction? \textbf{(d)} Is recoverability necessary, and which
design choices matter? \textbf{(e)} Do the resulting data improve
recovery-relevant dynamics?

\begin{figure}[t]
  \centering
  \includegraphics[width=1\linewidth]{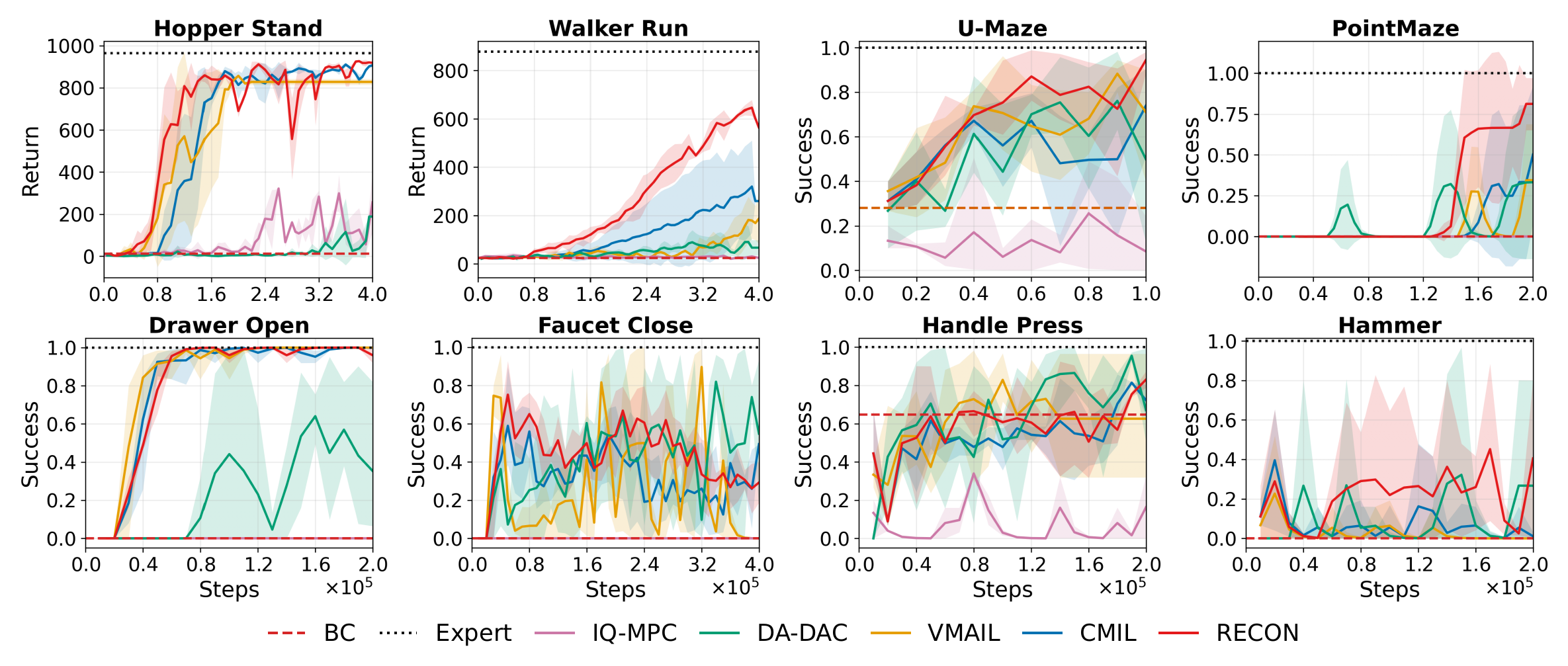}
  \caption{\textbf{Main experiments.} Imitation learning across eight environments under matched
  demonstrations and real-interaction budgets. RECON changes training-time
  acquisition; only its conservative main policy is evaluated.}
  \label{fig:main_curves}
\end{figure}

\paragraph{Protocol and baselines.}
We conduct experiments across continuous control, navigation, and manipulation tasks, including DMControl Hopper Stand and Walker Run, U-Maze, PointMaze, Meta-World Drawer Open, Faucet Close, Handle Press and Hammer. Rewards and
success labels are used only for evaluation. All methods share the same expert
demonstrations, interaction budgets, and 3 seeds. We compare with BC, which directly fits expert state--action pairs; data-augmented
replay-based DA-DAC~\citep{kostrikov2018discriminator}, which performs model-free adversarial
imitation with off-policy data reuse; V-MAIL~\citep{rafailov2021visual}, which performs
adversarial imitation on latent world-model rollouts; CMIL~\citep{kolev2024efficient},
which regularizes imagined policy optimization with epistemic uncertainty; and
IQ-MPC~\citep{li2025reward}, which combines reward-free world-model learning with
inverse soft-Q learning and latent model-predictive control.
RECON and CMIL use the same deployed policy objective and differ in online data
acquisition. Curves show aggregate performance with seed dispersion; complete
task details are in Appendix~\ref{app:environments} and baseline
implementations in Appendix~\ref{app:baselines}.
\vspace{-2mm}
\paragraph{\textbf{(a) Does targeted acquisition improve imitation?}}
Across all eight environments in Figure~\ref{fig:main_curves}, RECON achieves either the best or
competitive performance. The gains are particularly pronounced on Walker Run,
PointMaze, and Hammer, where effective imitation requires recovering from
deviations beyond the nominal expert trajectories. RECON also outperforms the most
closely related CMIL baseline on Faucet Close and Handle Press, since RECON and CMIL use the same deployed
actor and conservative policy objective, their comparison isolates the effect
of model acquisition from conservative model use itself.

\paragraph{\textbf{(b) Does targeted acquisition improve robustness beyond nominal training?}}
Table~\ref{tab:robustness} evaluates the main policy under action noise, delay,
impulses, and dynamics shifts unseen during training (definitions in
Appendix~\ref{app:robustness_protocol}). RECON improves
perturbation performance across the evaluated tasks, with especially
large gains under delay and altered dynamics. The results indicate that targeted acquisition improves the
world model beyond the nominal demonstration distribution by collecting
transitions around recoverable deviations, expanding the trust region around expert trajectories.

\vspace{-4mm}
\begin{table}[H]
\caption{Robust deployment of the main policy, reported as CMIL/RECON means over three seeds
(absolute $\Delta$). DMControl entries and deltas are returns; Maze and Meta-World entries are success percentages and deltas are percentage points
(pp). Higher is better.}
\label{tab:robustness}
\centering
\small
\setlength{\tabcolsep}{5pt}
\renewcommand{\arraystretch}{1}
\begin{tabular}{lrrrr}
\toprule
Environment & Clean & Noise & Delay & Impulse/scale\\
\midrule
Hopper & 905.8/\textbf{909.1}\retgain{3.3} & 862.9/\textbf{882.2}\retgain{19.3} & 408.5/\textbf{596.2}\retgain{187.7} & 864.0/\textbf{916.0}\retgain{52.0}\\
Walker & 421.2/\textbf{593.9}\retgain{172.7} & 401.6/\textbf{572.8}\retgain{171.2} & 133.0/\textbf{184.9}\retgain{51.9} & 378.6/\textbf{514.6}\retgain{136.0}\\
Drawer & \textbf{100.0}/\textbf{100.0}\ppsame & \textbf{100.0}/96.7\pploss{3.3} & 95.0/\textbf{96.7}\ppgain{1.7} & \textbf{100.0}/92.6\pploss{7.4}\\
Faucet & 30.0/\textbf{70.0}\ppgain{40.0} & 15.0/\textbf{75.0}\ppgain{60.0} & 5.0/\textbf{85.0}\ppgain{80.0} & 15.6/\textbf{66.1}\ppgain{50.5}\\
Handle & \textbf{75.0}/70.0\pploss{5.0} & 67.5/\textbf{77.5}\ppgain{10.0} & \textbf{80.0}/\textbf{80.0}\ppsame & 77.8/\textbf{86.1}\ppgain{8.3}\\
Hammer & 0.0/\textbf{2.5}\ppgain{2.5} & \textbf{8.3}/6.3\pploss{2.1} & 8.3/\textbf{26.3}\ppgain{17.9} & \textbf{10.0}/3.9\pploss{6.1}\\
U-Maze & 40.0/\textbf{45.0}\ppgain{5.0} & 31.7/\textbf{38.3}\ppgain{6.7} & 36.7/\textbf{44.2}\ppgain{7.5} & 32.5/\textbf{38.3}\ppgain{5.8}\\
PointMaze & 41.3/\textbf{100.0}\ppgain{58.7} & 42.0/\textbf{99.0}\ppgain{57.0} & 30.3/\textbf{100.0}\ppgain{69.7} & 16.0/\textbf{47.0}\ppgain{31.0}\\
\bottomrule
\end{tabular}
\end{table}
\paragraph{\textbf{(c) Where does $G$ redirect interaction?}}
\begin{figure}[t]
  \centering
  \includegraphics[width=1\linewidth]{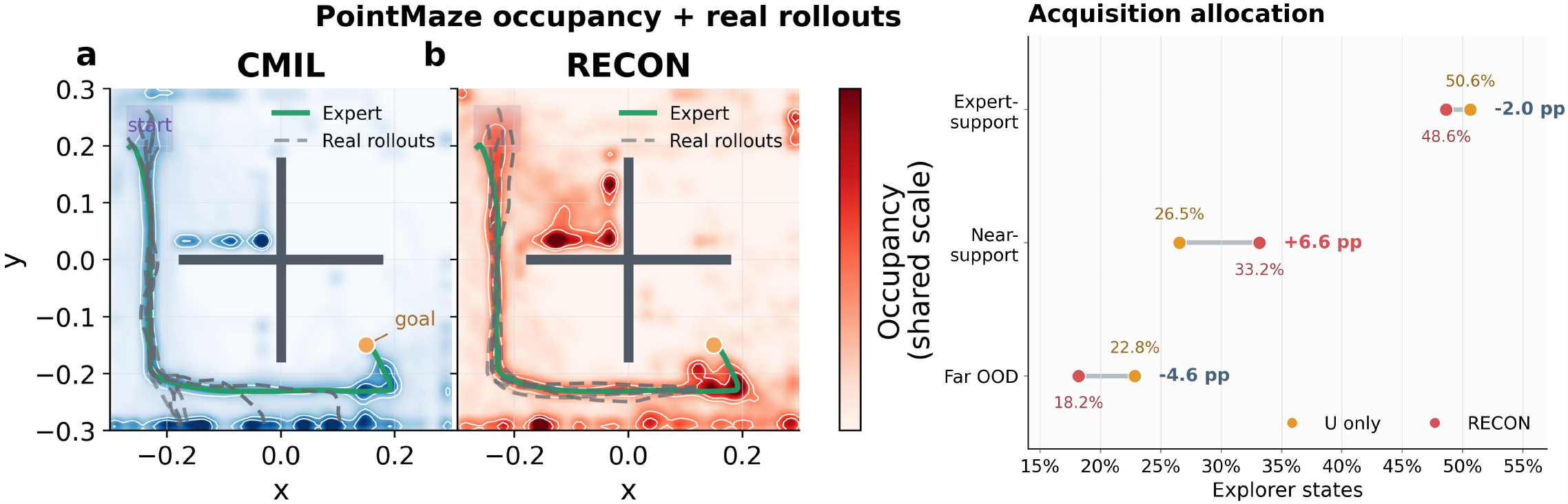}
  \caption{
\textbf{PointMaze visitation and acquisition.}
\textbf{Left:} State-visitation frequency of CMIL and RECON under the same
training budget, with expert trajectories and representative real rollouts overlaid.
\textbf{Right:} Acquisition allocation across expert-support, near-support,
and far-OOD regions, defined by distance to the expert state cloud.
}
  \label{fig:pointmaze_model}
\end{figure}

\begin{figure}[t]
  \centering
  \includegraphics[width=1\linewidth]{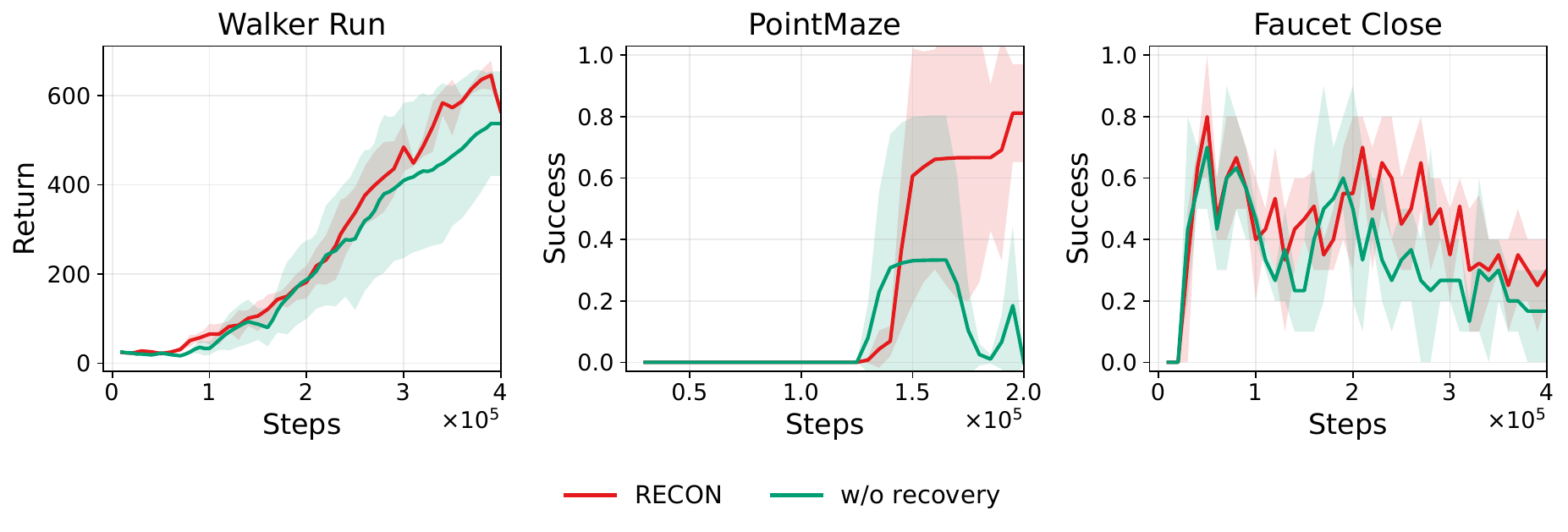}
  \caption{\textbf{Recoverability ablation.} ``w/o recovery'' removes the recoverability term and uses
$r_t^e=D_\psi(s_t,a_t)+\beta
U_\theta(s_t, a_t)$. Because recoverability gating rescales the effective uncertainty bonus in
RECON, we use $\beta=5$ for the ungated variant for a fair comparison.}
  \label{fig:recovery_ablation}
\end{figure}

Figure~\ref{fig:pointmaze_model} shows that around the lower
turn, CMIL's conservative acquisition provides limited coverage of recovery
dynamics near the expert trajectory, so inertia-induced deviations can drive the
policy into the wall, resulting in noticeably higher visitation density along the
boundary than RECON. To further characterize acquisition allocation, we partition explorer states
into expert-support, near-support, and far-OOD regions by distance to the
expert state cloud. RECON shifts interaction toward near-support states and
away from far-OOD novelty. Consistent with the acquisition blind spot in Theorem~\ref{thm:real_model_error}, this indicates that recoverability conditioning improves coverage of recovery-relevant dynamics that nominal conservative collection can leave underexplored.
\paragraph{\textbf{(d) Is recoverability necessary, and which choices matter?}}
Figure~\ref{fig:recovery_ablation} isolates the role of recoverability by removing $G$ while retaining the explorer's uncertainty bonus. The resulting uncertainty-only collector
can preserve early learning, but degrades later performance, most clearly on
Walker Run and PointMaze. Consistent with Theorem~\ref{thm:recovery_weighted_control}~\ref{thm:recovery_information_gain}, uncertainty alone identifies under-specified dynamics but not which uncertainty is useful for imitation; Appendix~\ref{1} Figure~\ref{fig:additional_diagnostics} (b) further shows that recoverability-gated acquisition yields the largest uncertainty reduction.

\begin{figure}[t]
  \centering
  \includegraphics[width=0.95\linewidth]{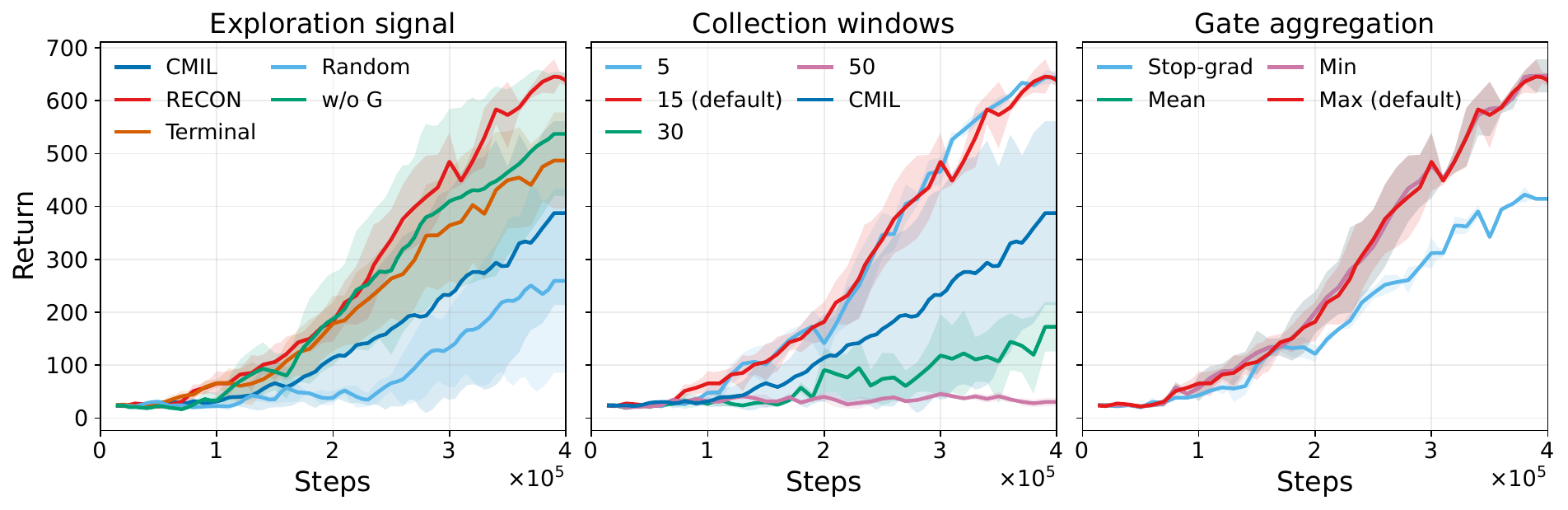}
  \caption{\textbf{Design ablation.} \textbf{Left:}
  alternative exploration signals. \textbf{Middle:} temporal placement of fixed-length
  collection bursts. \textbf{Right:} gate aggregation and the stop-gradient control.}
  \label{fig:walker_controls}
\end{figure}
Figure~\ref{fig:walker_controls} further separates the acquisition signal from the remaining design
choices. Random collection improves generic model coverage but spends too
little interaction on task-relevant deviations, while a terminal-only signal
captures expert compatibility without modeling whether the main policy can
actually recover through the dynamics. RECON instead evaluates recoverability
through multi-step main-policy imagination. With 10-step collection bursts, the
default 15-window schedule (totally 50-window) balances exploration coverage and main-policy control.
Retaining pathwise gradients through the recovery rollout clearly outperforms
stop-gradient, while the choice of tail aggregation has comparatively little
effect.

\paragraph{\textbf{(e) Does targeted acquisition improve the relevant model?}}
\begin{figure}[!t]
  \centering
  \includegraphics[width=1\linewidth]{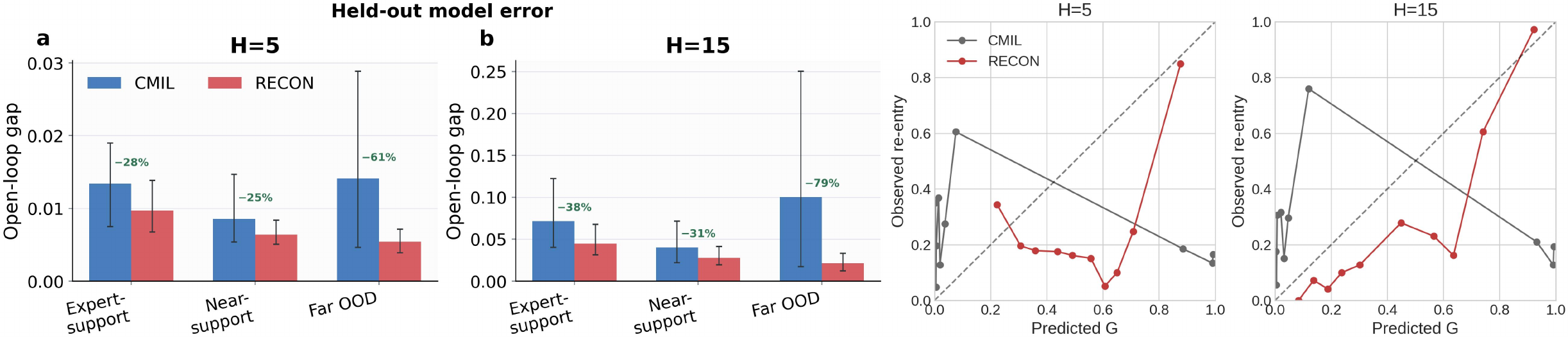}
  \caption{\textbf{Diagnostics of recoverability conditioning.} \textbf{Left:} Held-out discounted latent open-loop prediction gap across different regions. \textbf{Right:} Predicted recoverability $G_H$ (rollout horizon $H\in\{5,15\}$) versus empirical expert-tube
re-entry.}
  \label{fig:recovery_mechanism}
\end{figure}

As summarized in Table~\ref{tab:wm_diagnostics}, RECON generally reduces
world-model prediction error and ensemble uncertainty across the evaluated
tasks. Replay error may increase because RECON's replay contains more exploratory
trajectories, as in PointMaze. Figure~\ref{fig:recovery_mechanism} (left) shows lower held-out error across distance-based coverage regions.
Figure~\ref{fig:recovery_mechanism} (right) further validates the
acquisition signal itself: discriminator-based recoverability is strongly
correlated with empirical recoverability measured from real rollouts, and this
correspondence is markedly stronger for the policy trained with RECON than for
CMIL.

\begin{table}[!ht]
\vspace{-3mm}
\caption{World-model diagnostics, reported as CMIL/RECON (relative change). Gaps are latent open-loop
prediction errors defined in 
Appendix~\ref{app:error}. Lower is better; uncertainty is in $\times10^{-3}$.}
\label{tab:wm_diagnostics}
\centering
\small
\setlength{\intextsep}{1pt}
\begin{tabular}{lccc}
\toprule
Environment & Expert H15 gap & Replay H5 gap & Uncertainty\\
\midrule
Hopper & 1.229/\textbf{0.979}\gain{20.3} & 0.397/\textbf{0.272}\gain{31.5} & 2.8/2.8\unchanged\\
Walker & 0.608/\textbf{0.507}\gain{16.6} & 0.250/\textbf{0.214}\gain{14.4} & 8.3/\textbf{8.0}\gain{3.6}\\
Drawer & 1.537/\textbf{1.219}\gain{20.7} & 0.499/\textbf{0.394}\gain{21.0} & 4.0/\textbf{3.3}\gain{17.5}\\
Faucet & 0.774/\textbf{0.720}\gain{7.0} & 0.278/\textbf{0.272}\gain{2.2} & \textbf{3.1}/3.2\regress{3.2}\\
Handle & 1.072/\textbf{0.817}\gain{23.8} & 0.401/\textbf{0.330}\gain{17.7} & 3.1/\textbf{2.8}\gain{9.7}\\
Hammer & \textbf{0.933}/0.934\regress{0.2} & 0.349/\textbf{0.343}\gain{1.6} & 3.2/\textbf{2.9}\gain{8.3}\\
U-Maze & \textbf{7.049}/7.060\regress{0.2} & \textbf{2.450}/2.471\regress{0.9} & 8.1/\textbf{7.7}\gain{5.3}\\
PointMaze & 0.0535/\textbf{0.0331}\gain{38.1} & \textbf{0.0056}/0.0070\regress{25.0} & 11.9/\textbf{6.5}\gain{45.4}\\
\bottomrule
\end{tabular}
\end{table}
\section{Discussion and Limitations}
RECON improves the world model selectively rather than uniformly: targeted acquisition need only improve dynamics relevant to recovery from deviations around expert behavior. A current limitation is the additional model parameters and training computation introduced by the separate explorer. Future work may improve the two signals underlying acquisition by developing better calibrated epistemic uncertainty estimates and more reliable recoverability estimates beyond finite-horizon world-model imagination and discriminator scores, especially farther from current data support.

\section{Conclusion}
Conservative model use determines how an imperfect world model should guide policy optimization, but not where limited interaction should be spent to improve that model. RECON addresses this complementary acquisition problem by combining a conservative main policy with recoverability-conditioned exploration. Across control, navigation, and manipulation, this improves imitation, robustness and model accuracy, opening a complementary direction to conservative policy optimization: learning not only how to use an imperfect world model, but also where to improve it.

\subsection*{AI use statement}

In this work, we used generative AI tools to aid and polish writing, including improving wording, grammar, and clarity, and to assist with literature retrieval and discovery, such as identifying potentially relevant related work. All AI-assisted content was reviewed by the authors, and literature identified with AI assistance was checked against the corresponding original sources before inclusion. We take responsibility for the final content of this work, including all text, claims, citations, and artifacts produced with the aid of generative AI.

\subsection*{REPRODUCIBILITY STATEMENT}
Complete proofs of the theoretical results are provided in Appendix~\ref{app:proofs}. Appendix~\ref{app:experimental_details} specifies the training algorithm, model and policy hyperparameters, environments and observations, robustness perturbation protocol, world-model evaluation, and baseline implementations. All baselines are evaluated using their official implementations and the corresponding recommended hyperparameter settings. Additional diagnostic experiments are reported in Appendix~\ref{app:add}. The complete implementation of RECON, together with the code required to reproduce the reported experiments, will be released publicly upon acceptance.
\bibliography{iclr2027_conference}
\bibliographystyle{iclr2027_conference}

\clearpage
\appendix

\section{Detailed Proofs}
\label{app:proofs}

\subsection{Notation and regularity conditions}
Write $x=(s,a)$ and let $d_{M,q}^{\pi}$ denote the normalized discounted
state--action occupancy of policy $\pi$ under transition kernel $M$ and initial
distribution $q$.  Values are unnormalized discounted returns.  We assume
$0<\gamma<1$, a common measurable reward with $|r|\le R_{\max}$, and Markov
kernels on standard Borel spaces, so the conditional distributions below exist.
All equalities and inequalities between densities are understood almost
everywhere with respect to the indicated reference measure.

For the Bayesian statements, $p_k(\mathrm d\omega)=p(\mathrm d\omega\mid
\mathcal D_k)$ is the current posterior and
\begin{equation}
\widehat T_k(\mathrm dy\mid x)
=\int T_\omega(\mathrm dy\mid x)p_k(\mathrm d\omega)
\label{eq:posterior_predictive}
\end{equation}
is its posterior predictive kernel.  A prospective $Y_x$ is drawn by first
drawing $\omega\sim p_k$ and then $Y_x\sim T_\omega(\cdot\mid x)$.  Distinct
prospective observations are conditionally independent given $\omega$ and their
inputs.  The round-$k$ gate $g_k$ and reference distribution $\nu$ are fixed
while evaluating a query; this is essential because otherwise an additional
term accounts for changing the acquisition target itself.

\subsection{Proof of Theorem~\ref{thm:real_model_error}}
We first record the expectation--TV inequality used throughout.  Under the
convention $D_{\rm TV}(P,Q)=\sup_A|P(A)-Q(A)|$, every bounded measurable $f$
satisfies
\begin{equation}
|\mathbb E_Pf-\mathbb E_Qf|
\le 2\|f\|_\infty D_{\rm TV}(P,Q).
\label{eq:tv_expectation}
\end{equation}
This follows from the dual representation
$2D_{\rm TV}(P,Q)=\sup_{\|h\|_\infty\le1}|\mathbb E_Ph-\mathbb E_Qh|$.

For a fixed policy, define the state kernels
\begin{equation}
P^\pi(\mathrm ds'\mid s)=\int T(\mathrm ds'\mid s,a)\pi(\mathrm da\mid s),
\quad
\widehat P^\pi(\mathrm ds'\mid s)=\int \widehat T_{\mathcal D}
(\mathrm ds'\mid s,a)\pi(\mathrm da\mid s).
\end{equation}
Let $r^\pi(s)=\int r(s,a)\pi(\mathrm da\mid s)$.  The Bellman equations and
the resolvent identity imply
\begin{align}
V_T^\pi-V_{\widehat T}^\pi
&=(I-\gamma P^\pi)^{-1}r^\pi
 -(I-\gamma\widehat P^\pi)^{-1}r^\pi\\
&=\gamma(I-\gamma P^\pi)^{-1}
(P^\pi-\widehat P^\pi)V_{\widehat T}^\pi.
\label{eq:proof_resolvent}
\end{align}
Because $\|V_{\widehat T}^\pi\|_\infty\le R_{\max}/(1-\gamma)$,
Equation~\ref{eq:tv_expectation} gives
\begin{align}
\left|(P^\pi-\widehat P^\pi)V_{\widehat T}^\pi(s)\right|
&\le\int\pi(\mathrm da\mid s)
\left|\int V_{\widehat T}^\pi(s')
[T-\widehat T_{\mathcal D}](\mathrm ds'\mid s,a)\right|\\
&\le\frac{2R_{\max}}{1-\gamma}
\int e_{\mathcal D}(s,a)\pi(\mathrm da\mid s).
\label{eq:one_step_value_error}
\end{align}
Moreover,
$q(I-\gamma P^\pi)^{-1}=\sum_{t\ge0}\gamma^tq(P^\pi)^t$.
Integrating Equation~\ref{eq:proof_resolvent} against $q$, substituting
Equation~\ref{eq:one_step_value_error}, and using the definition of normalized
occupancy therefore yields
\begin{align}
|V_{T,q}^\pi-V_{\widehat T,q}^\pi|
&\le\frac{2\gamma R_{\max}}{1-\gamma}
\sum_{t\ge0}\gamma^t
\mathbb E[e_{\mathcal D}(s_t,a_t)]\\
&=\frac{2\gamma R_{\max}}{(1-\gamma)^2}
\mathbb E_{d_{T,q}^\pi}[e_{\mathcal D}],
\end{align}
which proves Equation~\ref{eq:real_model_error_bound}.  Notice that the
resolvent follows $P^\pi$, so the error is averaged under real visitation; using
the learned-model occupancy here would require a separate change-of-measure
argument.

For the imitation statement, insert the learned-model value of the learner:
\begin{align}
V_T^{\pi_E}-V_T^\pi
&=V_T^{\pi_E}-V_{\widehat T}^{\pi}
  +V_{\widehat T}^{\pi}-V_T^\pi. 
\label{eq:imitation_decomposition}
\end{align}
The second term is bounded by the result above.  The normalized occupancy
identity $V=(1-\gamma)^{-1}\mathbb E_{d}[r]$ and
Equation~\ref{eq:tv_expectation} give
\begin{equation}
V_T^{\pi_E}-V_{\widehat T}^{\pi}
\le\frac{2R_{\max}}{1-\gamma}
D_{\rm TV}(d_{T,\mu_0}^{\pi_E},d_{\widehat T,\mu_0}^{\pi}).
\end{equation}
Combining this inequality with Equation~\ref{eq:imitation_decomposition}
establishes Equation~\ref{eq:imitation_real_error_bound}.

It remains to justify the coverage claim.  Let
$d_0=d_{T,\mu_0}^{\pi_m}$ and $d_q=d_{T,q}^{\pi_m}$.  If $d_q\ll d_0$, then
for the nonnegative error $e_{\mathcal D}$,
\begin{align}
\varepsilon_q(\pi_m,\mathcal D)
&=\int e_{\mathcal D}(x)
\frac{\mathrm d d_q}{\mathrm d d_0}(x)\,d_0(\mathrm dx)\\
&\le\left\|\frac{\mathrm d d_q}{\mathrm d d_0}\right\|_\infty
\varepsilon_{\mu_0}(\pi_m,\mathcal D).
\end{align}
Thus the comparison yields a finite uniform factor whenever
$\|\mathrm d d_q/\mathrm d d_0\|_\infty<\infty$.
If $d_q\ll d_0$ but this essential supremum is unbounded,
the change-of-measure identity remains valid but does not provide
a finite uniform multiplicative bound.

Conversely, if $d_q\not\ll d_0$, there is a measurable $A$ with
$d_0(A)=0$ and $d_q(A)>0$.  The bounded error function $e=\mathrm{1}_A$ then
has zero nominal expectation and positive recovery expectation.  Equivalently,
on any nondegenerate successor space one may choose kernels that agree on
$A^c$ and have disjoint successor laws on $A$.  Hence no finite constant can
control recovery error from nominal error uniformly over transition kernels.

\subsection{Proof of Theorem~\ref{thm:recovery_weighted_control}}
For compactness define
\begin{equation}
e_{\omega,k}(x)=D_{\rm TV}(T_\omega(\cdot\mid x),
\widehat T_k(\cdot\mid x)),\qquad
\bar e_k(x)=\int e_{\omega,k}(x)p_k(\mathrm d\omega).
\end{equation}
Let $d_{\omega,q}=d_{T_\omega,q}^{\pi_m}$.  By the assumed occupancy
domination $d_{\omega,q}(x)\le C_q\nu(x)$ for posterior-almost every $\omega$,
nonnegativity and Tonelli's theorem give
\begin{align}
\mathbb E_{p_k}[\varepsilon_q(\omega,\mathcal D_k)]
&=\int p_k(\mathrm d\omega)\int e_{\omega,k}(x)
d_{\omega,q}(\mathrm dx)\\
&\le C_q\int\nu(\mathrm dx)\int e_{\omega,k}(x)p_k(\mathrm d\omega)
=C_q\mathbb E_\nu[\bar e_k].
\label{eq:posterior_occupancy_domination}
\end{align}
The identity $1=g_k+(1-g_k)$ separates the last expectation without dropping
either part:
\begin{equation}
\mathbb E_\nu[\bar e_k]
=\mathbb E_\nu[g_k\bar e_k]
 +\mathbb E_\nu[(1-g_k)\bar e_k].
\label{eq:proof_gate_split}
\end{equation}
Pinsker's inequality applied for each $(x,\omega)$, followed by Jensen's
inequality for the concave square root, yields
\begin{align}
\bar e_k(x)
&\le\int\sqrt{\tfrac12D_{\rm KL}
(T_\omega(\cdot\mid x)\Vert\widehat T_k(\cdot\mid x))}
p_k(\mathrm d\omega)\\
&\le\sqrt{\ell_{\mathcal D_k}(x)/2}.
\label{eq:posterior_pinsker}
\end{align}
Since $0\le g_k\le1$, write
$g_k\sqrt{\ell}=\sqrt{g_k}\sqrt{g_k\ell}$ and apply
Cauchy--Schwarz under $\nu$:
\begin{align}
\mathbb E_\nu[g_k\bar e_k]
&\le\frac{1}{\sqrt2}
\mathbb E_\nu[\sqrt{g_k}\sqrt{g_k\ell_{\mathcal D_k}}]\\
&\le\sqrt{\frac{\mathbb E_\nu[g_k]
\mathbb E_\nu[g_k\ell_{\mathcal D_k}]}{2}}
=\sqrt{Z_k\mathcal R_k(\mathcal D_k)/2}.
\label{eq:weighted_cauchy}
\end{align}
Substituting Equations~\ref{eq:proof_gate_split} and
\ref{eq:weighted_cauchy} into Equation~\ref{eq:posterior_occupancy_domination}
proves the theorem.  The result is also valid when $Z_k=0$, in which case the
weighted-risk term vanishes.  The residual term is unavoidable unless the gate
upper-bounds recovery occupancy everywhere; retaining it makes explicit that
the theorem motivates targeted, not global, model accuracy.

\subsection{Proof of Theorem~\ref{thm:recovery_information_gain}}
Fix inputs $x$ and $z$.  From Equation~\ref{eq:posterior_predictive}, the
conditional mutual-information identity for a mixture distribution is
\begin{align}
I(\omega;Y_z\mid z,\mathcal D_k)
&=\int p_k(\mathrm d\omega)
D_{\rm KL}\!\left(T_\omega(\cdot\mid z)
\Vert\widehat T_k(\cdot\mid z)\right)\\
&=\ell_{\mathcal D_k}(z).
\label{eq:kl_mutual_information}
\end{align}
After observing $Y_x$, Bayes' rule replaces $p_k$ by
$p(\omega\mid\mathcal D_k,x,Y_x)$.  Averaging the resulting predictive KL over
the as-yet-unobserved $Y_x$ gives
\begin{equation}
\mathbb E_{Y_x\mid x,\mathcal D_k}
[\ell_{\mathcal D'_{k,x}}(z)]
=I(\omega;Y_z\mid z,x,Y_x,\mathcal D_k).
\label{eq:posterior_kl_after_query}
\end{equation}
Let $C=(z,x,\mathcal D_k)$.  Expanding the same mutual information in two
orders gives
\begin{align}
I(\omega,Y_x;Y_z\mid C)
&=I(Y_x;Y_z\mid C)+I(\omega;Y_z\mid Y_x,C),\\
I(\omega,Y_x;Y_z\mid C)
&=I(\omega;Y_z\mid C)+I(Y_x;Y_z\mid\omega,C).
\end{align}
The last term is zero by conditional independence of prospective observations.
Consequently,
\begin{equation}
I(\omega;Y_z\mid z,\mathcal D_k)
-I(\omega;Y_z\mid z,x,Y_x,\mathcal D_k)
=I(Y_z;Y_x\mid z,x,\mathcal D_k).
\label{eq:mi_chain_difference}
\end{equation}
Multiplying by the fixed, nonnegative $g_k(z)$, integrating with respect to
$z\sim\nu$, and using Tonelli's theorem with
Equations~\ref{eq:kl_mutual_information}--\ref{eq:mi_chain_difference} proves
Equation~\ref{eq:recovery_information_gain}.

We finally make the cellwise specialization precise.  Suppose the input space
is a discrete partition into cells, the posterior factorizes over cell
parameters $\omega=(\omega_x)_x$, and an observation at $x$ depends only on
$\omega_x$.  Then $I(Y_z;Y_x\mid z,x,\mathcal D_k)=0$ for $z\ne x$, while for
$z=x$ a fresh replicate $Y_x'$ gives
\begin{equation}
\mathcal R_k(\mathcal D_k)-
\mathbb E[\mathcal R_k(\mathcal D'_{k,x})]
=\nu(x)g_k(x)I(Y_x';Y_x\mid x,\mathcal D_k).
\label{eq:stochastic_cell_gain}
\end{equation}
For stochastic transitions, the Markov chain
$Y_x-\omega_x-Y_x'$ implies that this predictive information is at most
$I(\omega_x;Y_x\mid x,\mathcal D_k)$.  If successors are deterministic given
$\omega_x$, then $Y_x=Y_x'$ almost surely and both quantities equal
$H(Y_x\mid x,\mathcal D_k)$, yielding the stated $\Gamma_k(x)$.  Thus generic
ensemble disagreement is a tractable proxy for the exact gain, not an equality
claimed for arbitrary stochastic neural dynamics.

\subsection{Latent-model specialization and pathwise gate gradient}
\label{app:gradient}
The analysis requires only a gate in $[0,1]$ held fixed during one acquisition
round; it does not rely on the discounted-average aggregation in
Equation~\ref{eq:round_recoverability}.  The implemented tail maximum in
Equation~\ref{eq:implemented_gate} is therefore a valid instantiation of the
same weighting principle.

Let $\phi$ parameterize a reparameterized explorer rollout
$\widehat s_{0:T}(\phi),\widehat a_{0:T-1}(\phi)$ and set
$U_t(\phi)=U_\theta(\widehat s_t,\widehat a_t)$.  Starting from its endpoint
$\widetilde s_0=\widehat s_T(\phi)$, the frozen main policy and world model
produce a deterministic recovery rollout and gate $G(\widehat s_T(\phi))$.
For fixed nonnegative return weights $w_t$, the acquisition component is
\begin{equation}
A(\phi)=\mathbb E_\epsilon\left[
\sum_{t=0}^{T-1}w_tU_t(\phi)G(\widehat s_T(\phi))\right],
\end{equation}
where $\epsilon$ collects actor and latent reparameterization noise.  Under the
usual dominated-differentiation condition, the pathwise gradient is
\begin{align}
\nabla_\phi A
=\mathbb E_\epsilon\sum_{t=0}^{T-1}w_t\bigg[
&G(\widehat s_T)\nabla_\phi U_t
+U_t\nabla_{\widetilde s_0}G(\widetilde s_0)
\frac{\partial\widehat s_T}{\partial\phi}\bigg].
\label{eq:recovery_gradient}
\end{align}
The first term moves the explorer toward epistemically informative transitions.
The second differentiates through the subsequent main-policy rollout and moves
the explorer endpoint toward states from which that fixed policy produces a
large recovery score.  Freezing model, discriminator, and main-policy
\emph{parameters} sets their parameter gradients to zero but preserves their
input Jacobians, which is exactly what the second term requires.

For the implemented $G=\max_{h\in\mathcal I}p_h$, let $h^*$ be its unique
maximizer.  Away from the measure-zero set of ties,
\begin{equation}
\nabla_{\widetilde s_0}G
=\frac{p_{h^*}(1-p_{h^*})}{\tau}
\nabla_{\widetilde s_0}
D_\psi(\widetilde s_{h^*},\widetilde a_{h^*}),
\label{eq:max_gate_gradient}
\end{equation}
where the final derivative includes all recurrent model and main-actor
Jacobians along the $h^*$-step recovery rollout.  At a tie, automatic
differentiation selects a valid max subgradient.  Applying stop-gradient to $G$
removes the second term of Equation~\ref{eq:recovery_gradient} while leaving the
uncertainty-gradient term intact, matching the control in
Figure~\ref{fig:walker_controls}. Our experiments use the continuous Gaussian
RSSM configuration, so its reparameterized samples preserve this pathwise
gradient.

Finally, suppose the latent belief is sufficient, and learned models
use the same observation kernel $K(\mathrm do'\mid s')$.  Their observable
one-step laws are the compositions $T_SK$ and $\widehat T_SK$.  The
data-processing inequality for every $f$-divergence gives
\begin{equation}
D_f(T_SK\Vert\widehat T_SK)
\le D_f(T_S\Vert\widehat T_S).
\label{eq:latent_data_processing}
\end{equation}
Thus controlling predictive discrepancy in belief space controls the induced
observable discrepancy under this sufficiency assumption.  It does not assert
that ensemble standard deviation equals KL: the implemented $U_\theta$ is a
monotone epistemic surrogate, exact only under additional ensemble likelihood
assumptions.
\section{Implementation and Experimental Details}
\label{app:experimental_details}
\subsection{DreamerV2 backbone and actor--critic optimization}
\label{app:dreamer_backbone}
RECON uses the DreamerV2 training architecture with CMIL's continuous ensemble
RSSM option. An image encoder supplies the posterior state
$s_t=(h_t,z_t)$ in Equation~\ref{eq:rssm}; a recurrent prior predicts imagined
states, and a decoder reconstructs observations. The implementation sets
\texttt{discrete=false}: $z_t$ is a reparameterized Gaussian rather than the
categorical latent used in the original Atari configuration. Ten bootstrapped
prior heads share the deterministic state and expose their predicted means for
$U_\theta$ in Equation~\ref{eq:implemented_uncertainty}. Expert and online
sequences train the shared representation and world model, while replay
posteriors seed 15-step prior rollouts for behavior learning.

For policy $i\in\{main,explore\}$, let $r_t^i$ be the corresponding reward in
Equation~\ref{eq:dual_rewards}. Main and explorer each have an actor and two
critics; the actor uses their pointwise minimum
$V_t^i=\min_{k\in\{1,2\}}Q_{i,k}(\widehat s_t^i,\widehat a_t^i)$ and the
pathwise return
\begin{equation}
 R_t^{\lambda,i}=r_t^i+\gamma_t\!\left[(1-\lambda)V_{t+1}^i
 +\lambda R_{t+1}^{\lambda,i}\right],\qquad
 w_t=\prod_{j<t}\gamma_j .
 \label{eq:lambda_return}
\end{equation}
The DreamerV2 actor objective is regularized by expert behavior cloning,
\begin{align}
 J_i&=\mathbb E_{\widehat\rho_{\pi_i}}\sum_t w_t
 \left[(1-\lambda)V_t^i+\lambda R_t^{\lambda,i}
 +\eta_i\mathcal H(\pi_i(\cdot\mid\widehat s_t^i))\right],\nonumber\\
 \mathcal L_{\pi_i}&=-J_i+\kappa_i
 \mathbb E_{(s,a)\sim\mathcal D_E}[-\log\pi_i(a\mid s)],\nonumber\\
 \mathcal L_{Q_i}&=-\frac{1}{2}\sum_{k=1}^2
 \mathbb E_{\widehat\rho_{\pi_i}}\sum_t w_t
 \log p_{Q_{i,k}}\!\left(\operatorname{sg}[\overline R_t^{\lambda,i}]
 \mid\widehat s_t^i,\widehat a_t^i\right).
 \label{eq:common_actor}
\end{align}
Here $\overline R^{\lambda,i}$ replaces $V^i$ in
Equation~\ref{eq:lambda_return} by one uniformly sampled slow-target critic
head. The main critics additionally fit one-step replay targets. Thus only the
rewards and collected state distribution differ; the latent optimization
machinery is shared.

\subsection{Complete training algorithm}
\label{app:algorithm}
Algorithm~\ref{alg:recon} summarizes the complete RECON training loop.
The main policy uses uncertainty conservatively, whereas the explorer uses
recoverability-gated uncertainty to acquire real transitions for the shared
world model.
\begin{algorithm}[t]
\small
\caption{RECON training and data acquisition}
\label{alg:recon}
\begin{algorithmic}[1]
\Require Expert replay $\mathcal D_E$, interaction budget $B$
\Ensure Deterministic main policy $\pi_m$

\State Initialize world model $p_\theta$, discriminator $D_\psi$,
main policy $\pi_m$, explorer $\pi_e$, and online replay $\mathcal D$
\State Pretrain $p_\theta,\pi_m,\pi_e$ on $\mathcal D_E$
\State Collect $B_0$ random transitions into $\mathcal D$

\While{interaction budget is not exhausted}
    \State Sample stratified explorer windows $\mathcal W$
    \For{each environment step $t$}
        \State Select
        \[
        \pi_t=
        \begin{cases}
        \pi_e,& t\in\mathcal W,\\
        \pi_m,& \text{otherwise},
        \end{cases}
        \]
        execute $a_t\sim\pi_t(\cdot\mid s_t)$, and store the transition in $\mathcal D$

        \If{an update is scheduled}
            \State Sample $\xi_E\sim\mathcal D_E$ and $\xi\sim\mathcal D$
            \State Update the shared world model using
            $\mathcal L_{\rm wm}(\xi_E\cup\xi)$

            \State Imagine $\widehat\tau_m$ with $\pi_m$ and update $D_\psi$
            \State Update $\pi_m$ using
            \[
            r_t^m
            =
            D_\psi(\widehat s_t^m,\widehat a_t^m)
            -\alpha U_\theta(\widehat s_t^m,\widehat a_t^m)
            \]

            \State Imagine $\widehat\tau_e$ with $\pi_e$ and set
            $\widetilde s_0=\widehat s_H^e$
            \State Roll out the frozen main policy for $H_r$ steps:
            \[
            \widetilde a_h=\operatorname{mode}\pi_m(\cdot\mid\widetilde s_h),
            \qquad
            \widetilde s_{h+1}=p_\theta(\cdot\mid\widetilde s_h,\widetilde a_h)
            \]

            \State Compute recoverability
            \[
            G_T=
            \max_{h=H_r-L+1:H_r}
            \sigma\!\left(
            \frac{D_\psi(\widetilde s_h,\widetilde a_h)-\delta_k}{\tau}
            \right)
            \]

            \State Update $\pi_e$ using
            \[
            r_t^e
            =
            D_\psi(\widehat s_t^e,\widehat a_t^e)
            +\beta U_\theta(\widehat s_t^e,\widehat a_t^e)G_T
            \]
            while freezing $\theta$, $\psi$, and $\pi_m$
        \EndIf
    \EndFor
\EndWhile

\State \Return $\operatorname{mode}\pi_m$
\end{algorithmic}
\end{algorithm}

\subsection{Hyperparameters}
\label{app:hyperparameters}
Tables~\ref{tab:common_hyperparameters}
reports the full configuration of RECON method; these values are shared across environments and seeds. Environment rewards and
success labels were not used for training.

\begin{table}[t]
\caption{Shared RECON hyperparameters.}
\label{tab:common_hyperparameters}
\centering
\small
\setlength{\tabcolsep}{4.2pt}
\begin{tabular}{lll}
\toprule
Component & Hyperparameter & Value \\
\midrule
Input/replay & observation; replay capacity & $64\!\times\!64$ RGB; $5\times10^6$ transitions \\
& sequence batch; sequence length & 32; 50 \\
& random seed transitions; update ratio & 5,000; 2 updates per 5 actions \\
World model & RSSM deterministic/stochastic size & 400 / 360 (Gaussian) \\
& transition ensemble; shared hidden size & 10 heads; 400 \\
& encoder/decoder CNN depth & 48 / 48 \\
& encoder kernels; decoder kernels & $(4,4,4,4)$; $(5,5,6,6)$ \\
& KL scale; free nats; balance & 1.0; 1.0; 0.8 \\
& model optimizer & Adam, lr $3\times10^{-4}$, $\epsilon=10^{-5}$ \\
& weight decay; gradient clip & $10^{-6}$; 100 \\
Behavior & actor/critic architecture & 4 layers $\times$ 400 ELU; two critics \\
& actor/critic optimizer & Adam, lr $3\times10^{-4}$ \\
& imagination horizon & 15 \\
& discount $\gamma$; lambda $\lambda$ & 0.99; 0.95 \\
& entropy coefficient & $10^{-4}$ \\
& target update period/fraction & 100 / 1.0 \\
& BC pretraining updates; main BC coefficient & 100 / 10 \\
Discriminator & architecture & 2 layers $\times$ 200 ELU, binary output \\
& optimizer; input noise std. & Adam, lr $3\times10^{-5}$; 2.5 \\
Main policy & uncertainty penalty $\alpha$ & 10 \\
Explorer & uncertainty coefficient $\beta$ & 10 \\
& imitation coefficient; explorer BC coefficient & 1; 10 \\
& collection schedule & 15 strata, one 10-step window per stratum \\
Recovery gate & rollout horizon $H_r$; tail length $L$ & 5; 3 \\
& aggregation; temperature $\tau$ & maximum; 0.05 \\
Optimization & behavior/model pretraining updates & 1,000 \\
& numerical precision; deterministic mode & FP32; enabled \\
\bottomrule
\end{tabular}
\end{table}

\subsection{Environments and expert observations}
\label{app:environments}
\begin{figure}[t]
  \centering
  \includegraphics[width=0.99\linewidth]{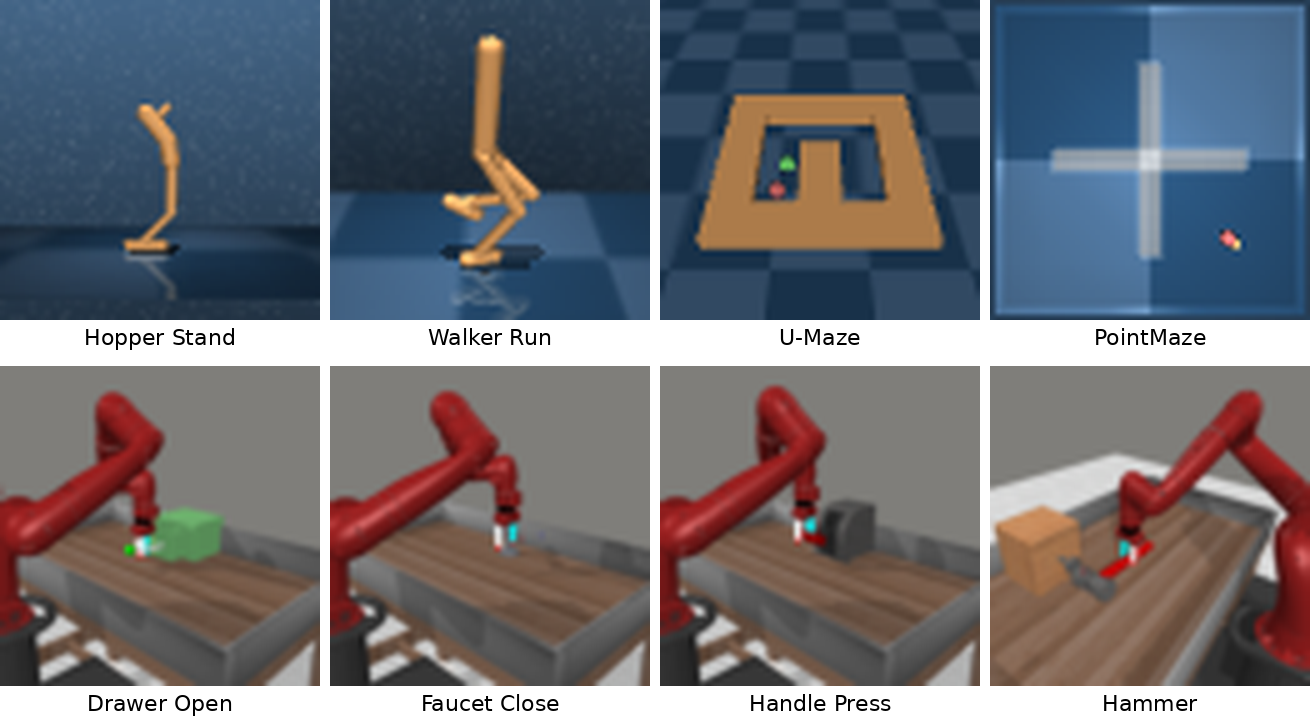}
  \caption{The eight evaluation environments.
  Policies receive the underlying $64\times64$ RGB observations.}
  \label{fig:environment_gallery}
\end{figure}

Figure~\ref{fig:environment_gallery} shows representative frames used in our experiments. It records the actual camera, rendering, and preprocessing
visible to every pixel-based method. DMC Hopper and Walker test locomotion and
contact dynamics; the two mazes test long-horizon correction around
bottlenecks; and the four Meta-World tasks test visually similar manipulation
scenes with distinct contact objectives.

For PointMaze diagnostics, let
$d_E(s)=\min_{s_E\in\mathcal S_E}\|p(s)-p(s_E)\|_2$,
where $p(s)$ is the planar environment position and
$\mathcal S_E$ denotes the expert state cloud.
We define expert-support by $d_E\le0.025$,
near-support by $0.025<d_E\le0.10$, and far-OOD by $d_E>0.10$.
These labels are used only for diagnostics and are unavailable to training.

\subsection{Robustness perturbation protocol}
\label{app:robustness_protocol}
For every method--environment pair, we evaluate three random seeds under the
same deployment configurations. These comprise \textbf{Clean:} Default
condition; \textbf{Noise}: two Gaussian action-noise levels,
$\sigma\in\{0.05,0.10\}$; \textbf{Delay:} two action delays,
$d\in\{1,2\}$ environment steps; and \textbf{Impulse/scale:} nine dynamics shifts including signed single-step action impulses of magnitude
$0.15$ or $0.30$ at $t=50$, component-wise action dropout with probability
$0.10$, action scaling by $0.85$, mass scaling by $1.25$ or $0.80$, friction
scaling by $1.50$ or $0.70$, and the joint mass/friction scale
$(1.25,1.50)$. All perturbed actions are clipped to the original action
bounds.

Each configuration is evaluated independently. We first average episodes
within a configuration, then configurations within each column of
Table~\ref{tab:robustness}, and finally the three seeds. DMC tasks use episode
return, while navigation and Meta-World tasks use success percentage. The
single-step impulses test recovery from a localized deviation, action dropout
and scaling test actuator mismatch, and mass/friction changes test persistent
dynamics shift.

\subsection{Open-loop model prediction gap.}
\label{app:error}
We evaluate world-model accuracy in latent space rather than by pixel
reconstruction error. For a held-out trajectory, let
$f_{t+h}^{\mathrm{post}}$ denote the posterior latent feature inferred using
the real observation at time $t+h$. Starting from the posterior state at time
$t$, we roll the world model open-loop for $H$ steps using only the recorded
actions $a_t,\ldots,a_{t+H-1}$, without conditioning on subsequent
observations, and denote the predicted latent feature at step $t+h$ by
$\widehat f_{t+h}$. We define the discounted $H$-step open-loop prediction
gap at starting point $t$ as
\begin{equation}
\mathcal E_H(t)
=
\sum_{h=1}^{H}
\gamma^{h-1}
\frac{1}{d_f}
\left\|
\widehat f_{t+h}
-
f_{t+h}^{\mathrm{post}}
\right\|_2^2,
\qquad
\gamma=0.99,
\label{eq:open_loop_gap}
\end{equation}
where $d_f$ is the latent feature dimension. Reported errors average
$\mathcal E_H(t)$ over the batch and all valid rollout starting points.

\subsection{Baseline implementation details}
\label{app:baselines}
All baselines use the same expert replay, image resolution, action bounds,
environment wrappers, evaluation seeds, and per-task real-interaction budget.
No baseline receives privileged state. All baselines are faithfully reproduced from their official
implementations.

\paragraph{Behavior cloning (BC).}
BC uses the same image encoder, continuous actor, and expert batches as the
model-based methods. The actor is optimized for 100 updates by maximum
likelihood on expert actions and then frozen. The world model may continue to
fit incoming images for matched logging, but neither those images nor
environment rewards update the BC actor; thus online data do not alter its
deployed behavior.

\paragraph{V-MAIL.}
V-MAIL learns a variational latent dynamics model and
performs adversarial imitation using on-policy rollouts generated inside the
learned model. The original implementation uses an earlier DreamerV1-style
continuous RSSM; for a stronger and fair comparison, we replace
it with the continuous DreamerV2 RSSM used by CMIL/RECON while preserving the
V-MAIL imitation objective. We otherwise match the discriminator, replay
batches, actor--critic, BC regularizer, and update ratio, set the uncertainty
penalty to zero, and use a single actor for both collection and evaluation.
This avoids confounding the comparison with differences in the world-model
backbone.

\paragraph{CMIL.}
CMIL is the single-policy version of our backbone. Its imagined reward is
$D_\psi(s,a)-10U_\theta(s,a)$, and that same conservative actor collects all
online transitions and is evaluated. It shares every entry in
Table~\ref{tab:common_hyperparameters} except for the absent explorer and
recovery gate. This is the closest comparison because RECON leaves CMIL's
deployed objective unchanged.

\paragraph{DA-DAC.}
DA-DAC is implemented as pixel DrQ-SAC with a four-layer, 32-channel CNN, a
50-dimensional feature projection, three stacked frames, two 1024-unit actor
and critic layers, twin critics, and automatic entropy tuning. It applies a
four-pixel replication-pad random crop twice per image and uses the V-MAIL
Appendix-C reward $\log D(s,a)$. Actor and critic learning rates are
$10^{-3}$, the temperature learning rate is $10^{-4}$, batch size is 128,
target $\tau=.01$, and actor and target networks update every two critic
updates. Expert transitions are kept in a separate positive replay and are
never inserted into the policy replay as discriminator negatives. DA-DAC has
no world model, uncertainty term, or BC regularizer.

\paragraph{IQ-MPC.}
IQ-MPC is run with the inverse soft-$Q$ objective and latent MPC. For a fair comparison with pixel inputs, it uses three stacked RGB frames and the TD-MPC2 encoder; we set
model size 5 and otherwise retain the released optimizer and planner: batch
size 256, learning rate $3\times10^{-4}$, latent size 512, five $Q$ heads,
planning horizon 3, six CEM iterations, 512 samples, 64 elites, and 24 policy
trajectories. Only the step budget, evaluation frequency, and demonstration
path are changed to match our protocol.

\subsection{Compute, runtime, and memory overhead}
\label{app:compute}
RECON shares the encoder, RSSM ensemble, decoder, discriminator, and replay
with CMIL. Its only persistent networks are one additional actor, two critics,
and their targets; the five-step recovery rollout reuses the shared model and
main actor. Table~\ref{tab:compute_overhead} reports an artifact-level audit of
archived Walker checkpoints and the final 50 logged throughput records.

\begin{table}[H]
\caption{Compute overhead relative to CMIL.}
\label{tab:compute_overhead}
\centering
\setlength{\tabcolsep}{6pt}
\begin{tabular}{lrrr}
\toprule
Quantity & CMIL & RECON & Difference \\
\midrule
World-model parameters & 17.38M & 17.38M & 0 \\
Total stored parameters & 21.52M & 25.46M & +3.94M (+18.3\%) \\
Explorer share of RECON parameters & --- & 15.5\% & --- \\
FP32 storage for added parameters & --- & 15.8 MB & +15.8 MB \\
Serialized checkpoint & 83 MiB & 98 MiB & +15 MiB \\
Logged FPS, last 50 records & $2.541\pm0.060$ & $2.430\pm0.148$ & $-4.4\%$ \\
Real interaction at evaluation & main actor & main actor & identical \\
Deployed policy class & main actor & main actor & identical \\
\bottomrule
\end{tabular}
\end{table}

\section{Additional Experiments}
\label{app:add}
\begin{figure}[H]
  \centering
  \includegraphics[width=0.99\linewidth]{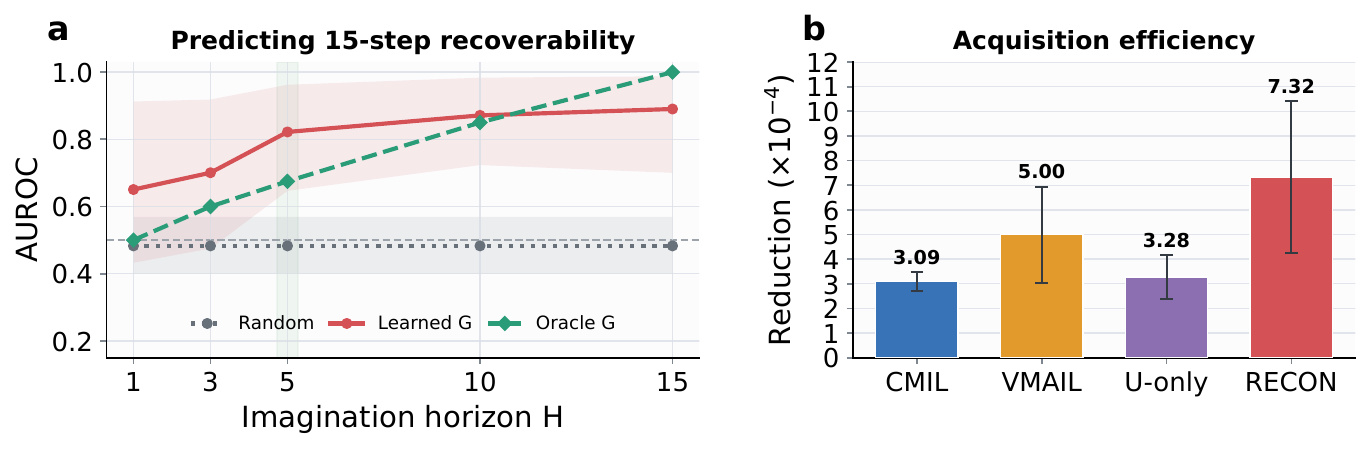}
  \caption{
  \textbf{Recoverability prediction and acquisition efficiency.}
  \textbf{(a)} AUROC between the predicted recoverability score and empirical
  recovery within 15 real-environment steps as the imagination horizon varies.
  \textbf{(b)} Reduction in ensemble uncertainty on fixed main-policy anchor
  transitions. The acquisition rewards are
CMIL: $D_\psi-10U_\theta$,
V-MAIL: $D_\psi$,
U-only: $D_\psi+5U_\theta$, and
RECON: $D_\psi+10U_\theta G$.
  }
  \label{fig:additional_diagnostics}
\end{figure}
\subsection{Recoverability prediction across imagination horizons.}
\label{1}
Figure~\ref{fig:additional_diagnostics}(a) evaluates recoverability prediction
on PointMaze. Empirical 15-step recoverability and the oracle predictor are
defined using true environment-space distance to the expert trajectory, whereas
the learned gate uses only discriminator scores along imagined main-policy
rollouts. We vary the imagination horizon $H$ and report AUROC against the
empirical recovery labels. Longer lookahead improves the learned predictor and
brings it closer to the oracle, supporting multi-step discriminator-based
imagination as a practical estimate of recoverability.

\subsection{Acquisition efficiency.}
\label{2}
Figure~\ref{fig:additional_diagnostics}(b) measures how efficiently online data
reduce ensemble uncertainty on a fixed set of main-policy anchor transitions.
For an anchor set $A$, we define the average ensemble uncertainty as
\begin{equation}
U_A(\theta)
=
\frac{1}{|A|}
\sum_{(s,a)\in A}
U_\theta(s,a),
\qquad
U_\theta(s,a)
=
\operatorname{MeanDim}
\left[
\operatorname{Std}_{m}
\bigl(\mu_{\theta,m}(s,a)\bigr)
\right].
\end{equation}
Keeping the anchor set fixed, the acquisition gain from model updates is
\begin{equation}
\Delta U_A
=
U_A(\theta_{\mathrm{before}})
-
U_A(\theta_{\mathrm{after}}),
\qquad
R
=
\frac{\Delta U_A}{\Delta N_{\mathrm{online}}},
\end{equation}
where $\Delta N_{\mathrm{online}}$ denotes the effective number of online
environment steps associated with the update interval.

RECON achieves the largest uncertainty reduction on the same anchor
transitions. In contrast, the ungated uncertainty-only collector is less
efficient than the reward-only V-MAIL collector, indicating that generic
uncertainty seeking can spend interaction on novel dynamics that contribute
little to reducing uncertainty around the main policy. Recoverability
conditioning directs exploration toward uncertain transitions that more
effectively improve the shared world model in regions relevant to deployment.

\subsection{Disagreement and held-out model error.}
\label{3}
\begin{figure}[!t]
  \centering
  \includegraphics[width=0.99\linewidth]{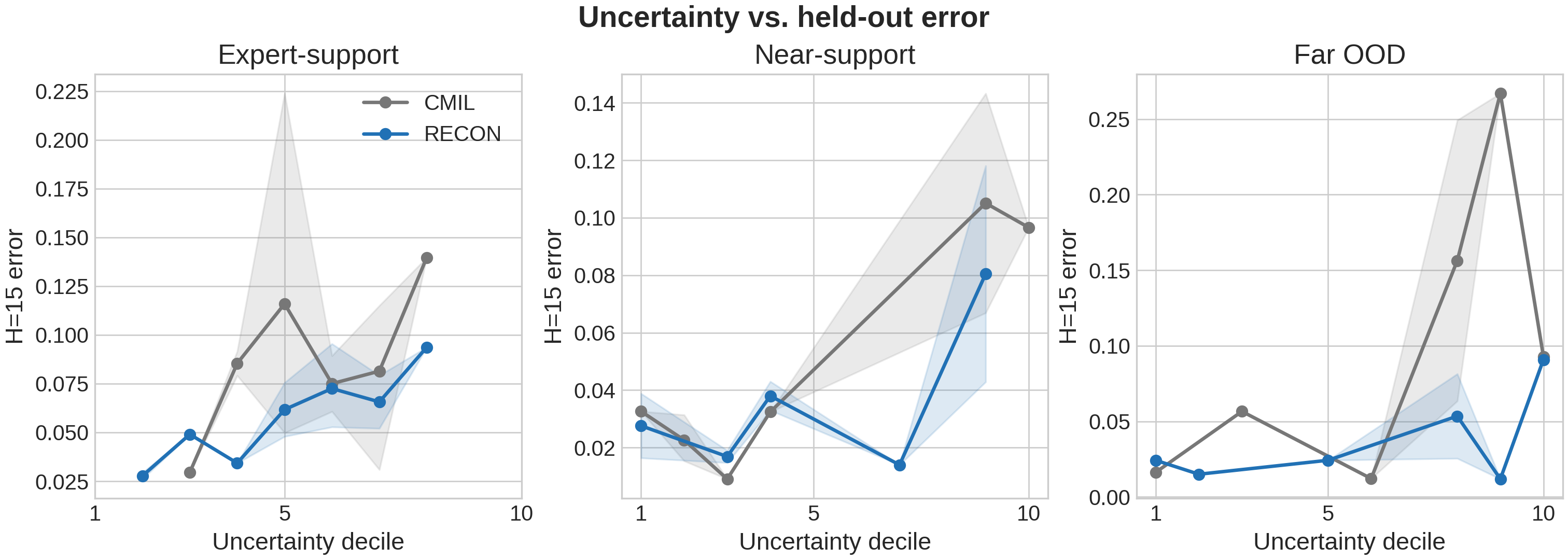}
  \caption{
  \textbf{Disagreement ranking and held-out model error.}
  $H=15$ open-loop prediction error across within-run uncertainty deciles on
  common held-out PointMaze spatial bins.
  }
  \label{fig:uncertainty_deciles}
\end{figure}

In Figure~\ref{fig:uncertainty_deciles}, within expert-support regions, larger ensemble disagreement generally
corresponds to larger held-out prediction error for RECON. The relationship is
weaker and non-monotone in near-support and far-OOD regions, where
long-horizon error additionally reflects compounding rollout error and sparse
data support. These results support using $U_\theta$ as an acquisition ranking
signal, without requiring it to provide a calibrated pointwise estimate of
model error.
\end{document}